\documentclass{article}
\usepackage{PRIMEarxiv}
\usepackage[numbers,sort&compress]{natbib}
\usepackage[utf8]{inputenc} 
\usepackage[T1]{fontenc}    
\usepackage{hyperref}       
\usepackage{url}            
\usepackage{booktabs}       
\usepackage{amsfonts}       
\usepackage{nicefrac}       
\usepackage{microtype}      
\usepackage{lipsum}
\usepackage{fancyhdr}       
\usepackage{graphicx}       
\usepackage{amsmath}
\graphicspath{{media/}}     
\usepackage[strings]{underscore}
\usepackage{colortbl} 
\usepackage{xcolor} 
\definecolor{LightBlue}{rgb}{0.68, 0.85, 0.9}
\definecolor{LightGreen}{rgb}{0.56, 0.93, 0.56}

\title{Improving Combined Detection and Classification of TEM Defects via Mask-Conditioned Latent Diffusion Augmentation
\thanks{\textit{\underline{Citation}}: 
\textbf{Authors. Title. Pages.... DOI:000000/11111.}} 
}
\author{
  Ni Li\textit{$^{1}$}, Nuohao Liu\textit{$^{1}$}, Ryan Jacobs\textit{$^{1}$}, Ajay Annamareddy\textit{$^{1}$}, Maciej P. Polak\textit{$^{1}$}, Kevin Field\textit{$^{2}$}, \\
   \textbf{Izabela Szlufarska\textit{$^{1}$}, Dane Morgan\textit{$^{1}$}} \\
  1. University of Wisconsin-Madison, Madison, WI\\
2. University of Michigan-Ann Arbor, Ann Arbor, MI\\
}

\begin{document}
\maketitle
\begin{abstract}
Analyzing microstructural defects in transmission electron microscopy (TEM) images, particularly in irradiated metal alloys, is often limited by the availability of high-quality, labeled data. To address this, we introduce a generative data augmentation approach using a mask-conditioned latent diffusion model (LDM) for synthesizing realistic TEM images with controllable, automatically labeled multi-class defect masks. Without requiring manual annotations for generation, our method enables the creation of synthetic image–mask pairs by sampling distributions learned from experimental masks. These generated data were used to augment small experimental datasets of varying sizes (10, 50, and 100 labeled experimental images) to train a Mask Regional Convolutional Neural Network (R-CNN) model for defect detection and classification. Our results show that generative augmentation yields small overall model performance improvements, with up to a 0.02 gain in the harmonic mean of detection and classification F1 scores (F1\textsubscript{HM}). However, we also find that the relative contributions to detection and classification improvement depend on the specific train/test data split. These findings highlight the potential of targeted generative models to enhance deep learning performance in data-scarce microscopy-based image quantification tasks.

\end{abstract}


\section{Introduction}

Irradiation from energetic particles introduces a broad spectrum of defect structures in metal alloys. The energetic particles can displace atoms, creating vacancies and interstitials that then evolve into a variety of extended defects such as dislocation lines, loops, and defect clusters.\cite{Was2016} These microstructural changes degrade the mechanical and physical properties of the material and must be accurately characterized to understand radiation damage mechanisms. Scanning Transmission Electron Microscopy (STEM) and Transmission Electron Microscopy (TEM) enable direct visualization of material defects at atomic to nanometer-scale resolution, making the two techniques indispensable for materials characterization, including characterization of radiation damage. STEM remains a primary tool for imaging these defects, and improving automated defect classification and identification in TEM images is critical for accelerated materials design and lifetime prediction in nuclear environments. However, the manual annotation of microscopy images is labor-intensive and requires significant domain expertise, posing a barrier to scaling automated analysis. 

Deep learning (DL) models have shown promise in automating feature labeling, with convolutional neural networks (CNNs) forming the backbone of object detection frameworks such as the Faster Regional CNN (R-CNN)\cite{ren2016faster}, Mask R-CNN\cite{he2018mask}, and You Only Look Once (YOLO)\cite{redmon2016look} models. These models have been successfully applied to detect a variety of structural features, including defects, nanoparticles, and atomic-scale structures in EM data.\cite{RN987,RN985,CHEN2023112073,RN988,Jacobs2023,RN984,RN986,RN989,RN990,Holm2020,DENNLER2021103069,OKTAY2019113,10960607,Baderot2022, Wu2025, TALLER2024100468, 10.1063/5.0274266,10.1063/5.0274266} Despite their success, the performance of these models still relies heavily on large, high-quality, and expertly annotated datasets, which are costly to produce, challenging to acquire, and where labeling objects of interest is subject to human bias\cite{10.1093/micmic/ozad067.767}. This limitation of painstakingly-acquired experimental data is especially severe in nuclear materials research, where the challenges of safely and correctly creating desired irradiation and thermal conditions make experimental data acquisition exceptionally challenging \cite{MORGAN2022100975}.

A common strategy to address microscopy data scarcity is synthetic dataset generation via physics-based simulations\cite{Lin2022,Lynch2025}. For example, Lynch et al. \cite{Lynch2025} combined experimentally acquired “clean” background images with computationally simulated defects based on physical models, producing realistic defect morphologies with pixel-perfect labels. Another approach used 3D modeling and rendering to create nanoparticle images with realistic SEM contrast, generating annotated datasets directly from the simulated geometries. \cite{CIDMEJIAS2021105958}. While effective, these methods require experts to have substantial domain expertise to determine which physical models are appropriate and how to apply them.
An alternative approach is to use generative models. While training latent diffusion models incurs substantially higher computational cost compared to conventional augmentation techniques such as physics-based synthetic image generation and MixUp-based strategies\cite{zhang2018mixup}, generative augmentation offers the ability to introduce new, physically realistic structural variations beyond simple perturbations of existing images. In practice, such approaches are expected to be useful when traditional augmentation methods do not yield object detection models with desired performance level s, or if acquisition of additional experimental data is particularly time consuming or costly. In addition to traditional and LDM generative augmentation approaches, another approach to augmenting experimental data is to generate synthetic data with physics-based approaches, such as that used in the work of Lynch et al. In that work, the authors created synthetic TEM images containing cavity defects. However, this approach required clean experimental backgrounds and a physics-based framework for simulating images of individual cavities. While the approach of Lynch et al. can be extended to handle other defect types, doing so is significantly harder than the latent diffusion model (LDM) approach used here, where extending to other defect types can be done by simply training the model with labeled instances of the new defect type. 

Both generative adversarial networks (GANs)\cite{GANs} and diffusion models\cite{diffusion} have gained attention for their ability to produce high-quality synthetic images. GANs have been widely adopted across computer vision tasks such as image classification and segmentation\cite{GAN-review,Ma2020}, and have also been extensively applied in microscopy imaging, particularly in medical and biological contexts\cite{GAN-med}. More recently, GAN-based approaches have also been explored for synthetic data generation in electron microscopy images within materials science applications.\cite{SHEN2024114407,yuan2024}. For example, tandem GAN pipelines have been applied to electron microscopy (EM) data to generate nanoparticle images and improve segmentation network performance in real-time in-situ experiments \cite{yuan2024}. However, such approaches have typically focused on single-object-class tasks, limiting their applicability to complex, multi-class segmentation problems.

Diffusion models, particularly LDMs\cite{LDM}, have demonstrated superior image quality and stability compared to GANs in applications ranging from medical imaging to EM image augmentation \cite{KAZEROUNI2023102846, Lu2024}. Prior work with conditional diffusion pipelines that generate both the segmentation masks and the corresponding images reports that augmenting the training set with these fully synthetic pairs can consistently improve downstream detection precision by a few percentage points, while the F1 score showed smaller gains.\cite{Mach2023}. These insights motivate our use of LDMs for the generation of synthetic TEM datasets for multi-class instance segmentation tasks. 

\begin{figure*}[ht]
 \centering
 \includegraphics[width=0.98\textwidth]{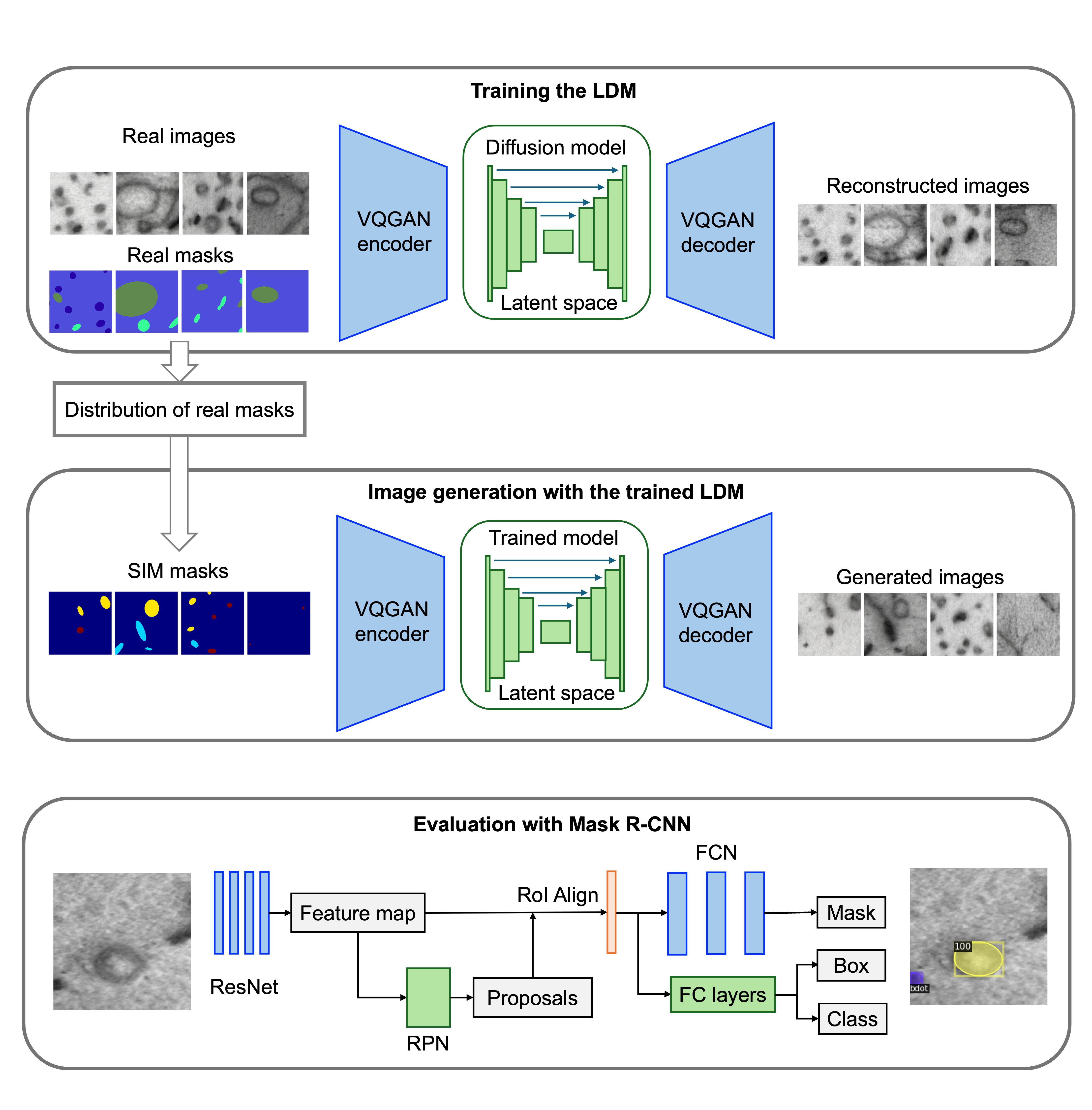}
\caption{Mask-conditioned latent diffusion pipeline for STEM image generation. Simulated (SIM) masks are sampled from the distributions of defect size, shape, and orientation from real defect masks and used to condition a latent diffusion model consisting of a VQGAN encoder-decoder and a U-Net in latent space. Mask R-CNN model was used to evaluate how well the generated images match the masks.}
 \label{fig:fig1}
\end{figure*}

In this work, we propose a mask-conditioned latent diffusion pipeline\cite{LDM} to generate automatically labeled synthetic STEM images aimed at improving multi-class instance segmentation and classification of irradiation-induced defects. Our approach uses simulated segmentation masks, generated by sampling statistical distributions of defect shape, size, orientation, and count from experimental data. The LDM model is then trained to synthesize realistic STEM images conditioned on these masks. As shown in Figure \ref{fig:fig1}, the proposed approach produces realistic image–mask pairs that effectively augment the limited experimental dataset. Throughout this work (including Figure \ref{fig:fig1}), we denote the generated masks as part of the image-mask pairs as “simulated masks” or SIM to differentiate our non-physics-based image generation process from the physics-based synthetic data generation in Lynch et al. We demonstrate that augmenting Mask R-CNN training with our synthetic data yields small but consistent improvements in overall model performance as measured by the harmonic mean of detection and classification F1 scores (as described in Section \ref{sec:evaluation_metrics})), compared to a model using only experimental data.

\section{Data and Methods}

The dataset used in this work is based on the on-zone STEM image collection compiled by Field et al.~\cite{FIELD2015746,FIELD201754,FIELD201720}, and was previously utilized by Jacobs et al.~\cite{RN985} for pixel-level defect segmentation using the Mask R-CNN framework. We adopt the same dataset, as well as the Mask R-CNN training and evaluation pipeline established in their study. The dataset comprises STEM images of irradiated FeCrAl alloys with varying compositions and microstructures imaged directly on the [100] zone axis (i.e., the "on-zone" imaging condition\cite{Parish2015}). Although the irradiation conditions (e.g., dose, temperature, etc.) differ across samples, all images were acquired under consistent imaging settings. The typical image resolution is $1024 \times 1024$ pixels. The primary defects of interest include black dots and two types of dislocation loops: $\langle100\rangle$ loops and $\langle111\rangle$ loops where the Miller indices indicate the loop Burgers vector. The $\langle100\rangle$ loops generally appear as single-edged (e.g., white-black-white contrast with the background, loop, and background) line features when viewed edge-on and can exhibit a characteristic double-beam contrast (e.g., white-black-white-black-white, where the central white contrast between the double-black contrast is considered the location of the dislocation core) when imaged face-on due to being in-plane (though these are less common). In comparison, $\langle111\rangle$ loops typically display a single-edged elliptical morphology with aspect ratios greater than one.\cite{YAO2013402,Parish2015} In addition to the STEM images, expert-annotated multi-class segmentation masks are provided for all images. We note that although images might also contain line dislocations, we ignore these features from the analysis as these are not labeled in the original dataset published by Jacobs et al.~\cite{RN985}.

\subsection{Data augmentation and splitting for training and testing}

We evaluated the impact of training data volume by selecting 10, 50, and 100 experimental images from the database (excluding the 22 test images) for model training. For each case, the same subset of experimental images was used in two ways: as training data for the latent diffusion model (LDM) to generate synthetic datasets, and as the baseline experimental-only training split for the Mask R-CNN model. To standardize the input size and increase dataset diversity, we applied a sliding window to crop the selected images and their corresponding masks into patches of size $256 \times 256$ pixels. Each cropped patch was further augmented through rotations ($90^\circ$, $180^\circ$, and $270^\circ$) and horizontal/vertical flipping. This process resulted in approximately 10{,}000 augmented patches per training split. All augmentation operations were applied identically to both images and masks to preserve alignment. To ensure a fair comparison across different training set sizes, we adopted a consistent augmentation target of 10{,}000 patches for all cases. This approach allows us to fully utilize the available experimental data under each condition. For LDM training, the augmented data was further split into 90\% for training and 10\% for validation.

\subsection{Training of mask-conditioned LDM}

With an autoencoder pretrained on ImageNet dataset fixed throughout, only the diffusion-denoising component of the latent diffusion model (LDM) was trained from scratch using the experimental STEM image data. The model was conditioned on segmentation-style masks, where each pixel encodes the defect type, enabling controllable generation of labeled structures. Training was performed until convergence, while avoiding overfitting, and the number of epochs required varied depending on the amount of available training data (10, 50, or 100 images). Training the LDM model took approximately 7 GPU-hours on a single NVIDIA V100, and generating 10{,}000 synthetic images took $\sim$8 GPU-hours (3 seconds per image), totaling $\sim$15 GPU-hours. Further training details, including selection of pretrained autoencoder model and loss curves, are provided in the Supplementary Material.

\subsection{Simulating masks based on real mask distribution}

To generate simulated segmentation masks (SIM masks) that resemble the characteristics of real experimental masks, we first analyzed the statistical distributions of several key properties of defects in the labeled experimental data. Each defect was approximated as an ellipse, and we extracted the distributions of the number of defects per image, the major axis length, minor axis length, and orientation angle of each defect type (i.e., $\langle100\rangle$ loop, $\langle111\rangle$ loop, and black dot). These distributions were then used to sample new masks that mimic the statistical properties of the experimental dataset. Specifically, we sampled the number of defects per image from the experimental distribution, then sampled the type, size, and orientation of each defect from their respective distributions. This process ensures that the synthetic masks preserve both the quantity and diversity of the real masks. The comparison between the real and simulated mask distributions in terms of the number of defects per image, the orientation angle and the size across the three defect types are shown in the Supplementary Material. The distributions show strong agreement between the simulated and real data. Overall, masks from ellipse-based sampling are well aligned with the experimental annotations, as human-labeled loop defects in the dataset are typically represented as elliptical structures. Under [100] on-zone imaging conditions, ⟨100⟩ loops project either edge-on as line features or in-plane with relatively weak contrast; in this dataset, in-plane loops represent only a small fraction (~5\%) of the total population. As a result, elliptical parameterization—including high-aspect-ratio masks—provides an effective and physically meaningful approximation of the labeled defect morphology, while allowing straightforward control over defect size, orientation, and count based on empirical distributions. One could also generate masks from some form of deep learning generative model, and such a choice might be much easier than a hand-tuned sampling in cases where the masks are highly complex shapes rather than simple ellipses. We explored generative approaches for mask synthesis in this work, but found it difficult to reproduce distributions similar to those in the experimental data. Specifically, LDM-generated masks often exhibited distortions and unintended connections between defect regions. The use of specialized generative models for structured segmentation outputs have the potential to further improve performance.\cite{Zargari2024Enhanced}

\subsection{Evaluation using Mask R-CNN model and comparison with the results of using the experimental dataset}
\label{sec:evaluation_metrics}

To evaluate the effectiveness of the generated data, we trained and tested a Mask R-CNN model under two different data conditions. In the baseline setting, the model was trained on 10,000 256$\times$256 pixels images derived from experimentally labeled data and evaluated on a separate test set of 22 1024 $\times$ 1024 pixel experimental images cropped into patches of size of 256$\times$256 pixels (this is called the "EXP" model). For comparison, a second model was trained on a combined dataset consisting of the same 10,000 256$\times$256 pixel experimental images and an additional 10,000 256$\times$256 pixel images generated by the LDM conditioned on simulated masks (this is called the "EXP+GEN" model). Both models were tested on the identical experimental test set to ensure consistency. To enable a fair comparison, the Mask R-CNN training configuration—including learning rate, optimizer, and other hyperparameters—was kept identical across both settings. Each model was trained for 100{,}000 iterations, and the final checkpoint was used to evaluate performance on the test split. We compared the results of the EXP and EXP+GEN models in terms of defect detection F1 score and defect classification F1 score to assess the impact of including synthetic data. These F1 scores were computed by aggregating predictions across all test images, treating the entire dataset as a whole. Specifically, true positives, false positives, and false negatives were summed across all defect instances in all test images, and then used to compute overall F1 scores. To enable a consistent and meaningful comparison across different experiments, we introduce harmonic F1 score termed {F1\textsubscript{HM}}, defined as the harmonic mean of detection F1 and classification F1:

\[
\text{F1}_{\text{HM}} = \frac{2 \cdot \text{F1}_{\text{detect}} \cdot \text{F1}_{\text{class}}}{\text{F1}_{\text{detect}} + \text{F1}_{\text{class}}}
\]

{F1\textsubscript{HM}} penalizes large discrepancies between detection and classification performance, ensuring that high scores in both are needed for a strong overall score. 

\section{Results and Discussion}
\subsection{Comparison of generated images with real images}

\begin{figure*}[ht]
 \centering
 \includegraphics[width=0.98\textwidth]{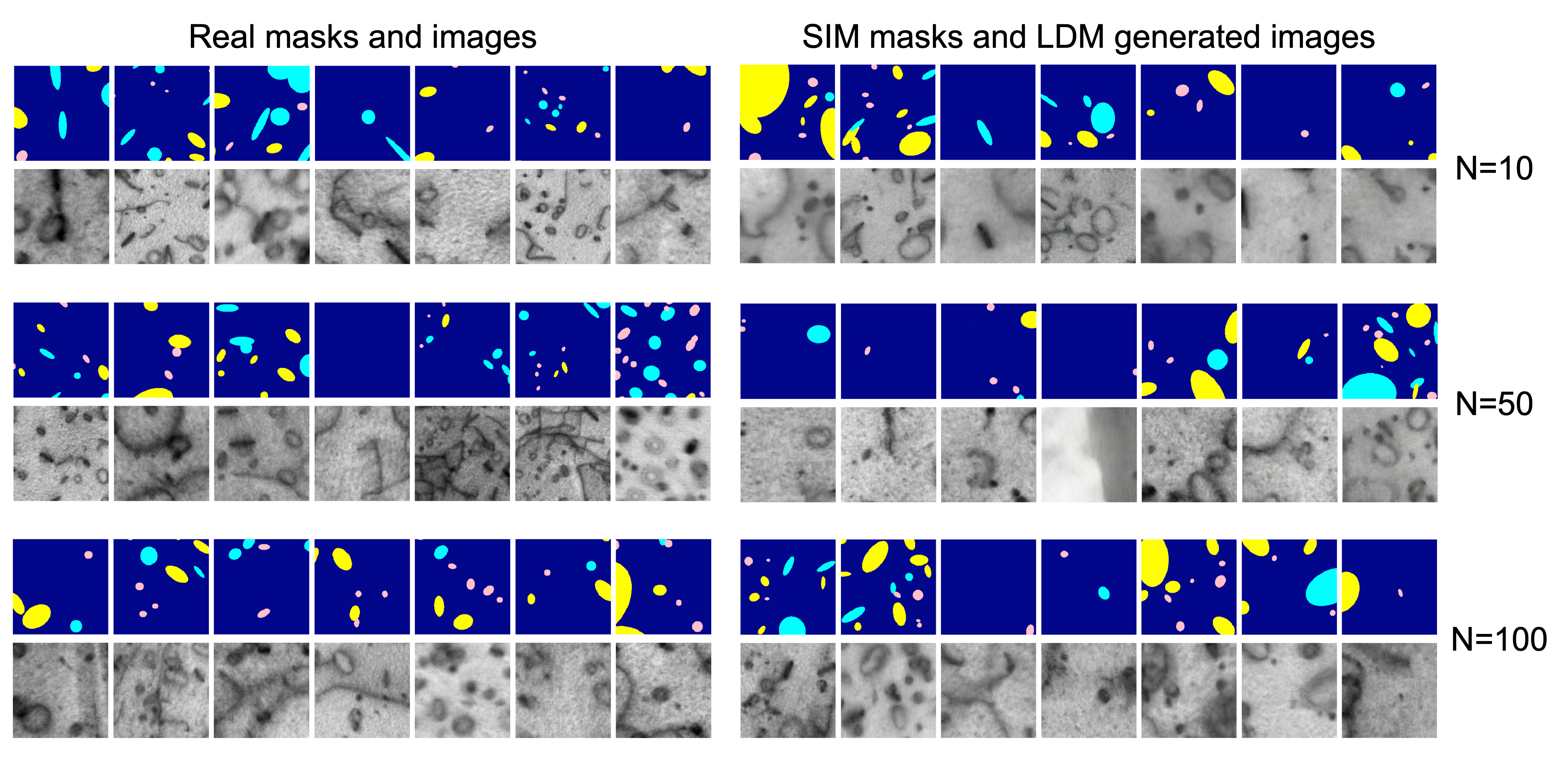}
\caption{Visual comparison between real (left) and generated data (right). Left: Real segmentation masks and corresponding experimental TEM images. Right: Simulated (SIM) masks sampled from real mask statistics and their corresponding LDM-generated images. From top to bottom, the generated samples correspond to models trained with 10, 50, and 100 experimental images, respectively.}
 \label{fig:compare}
\end{figure*}

Figure \ref{fig:compare} presents a visual comparison between the real and simulated masks and their corresponding STEM images. The left panel shows representative examples of real semantic segmentation masks and experimental STEM images. The right panel displays simulated masks, which were sampled from the distribution of real masks, along with synthetic STEM images generated by the mask-conditioned LDM. Each segmentation mask includes three defect types—$\langle100\rangle$ loops  (blue), $\langle111\rangle$ loops (yellow), and black dots (pink).  

The three rows on the right correspond to models trained with increasing amounts of experimental data: 10, 50, and 100 images, respectively. Visually, the LDM-generated images improve as the number of training images increases. The first row (trained on 10 images) shows blurrier features and more uniform backgrounds, while the second and third rows (50 and 100 images) yield sharper textures and more realistic defect morphology. This progression demonstrates that although the LDM is capable of producing visually consistent, label-consistent outputs, the model benefits from greater training data to capture the complexity of real STEM backgrounds and defect structures. Similar behavior has been observed in diffusion models trained on other image datasets, where images generated from models trained on small datasets are coarse.\cite{zhu2023fewshotimagegenerationdiffusion} Notably, because the training dataset consists of on-zone STEM images, diffraction-specific features such as coffee-bean contrast and the associated line of no contrast are not present and therefore cannot be evaluated. While the LDM reproduces the general morphology of in-plane loops, it does not consistently retain contrast variations along the full projected loop length, likely due to the absence of physics-based constraints in the model and the limited representation (~5\%) of such features in the training set.

\subsection{Quantitative evaluation of generated images from Mask R-CNN}

\begin{figure*}[ht]
 \centering
 \includegraphics[width=0.98\textwidth]{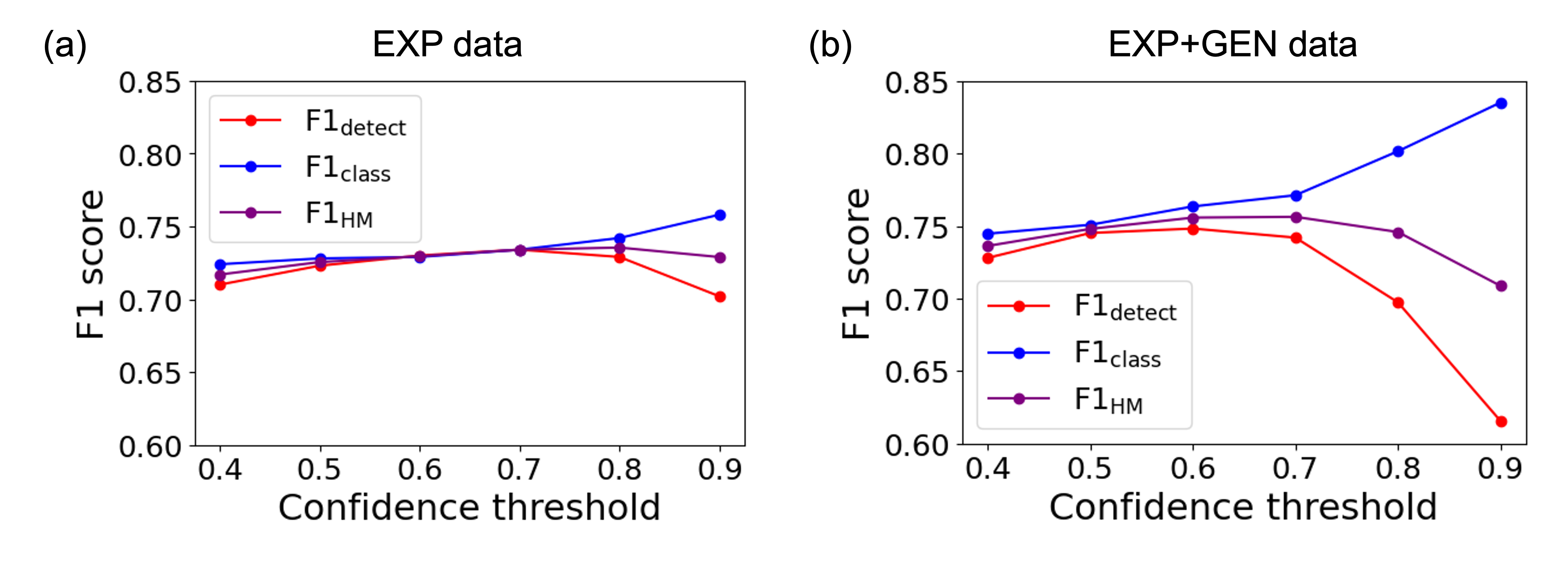}
\caption{Detection F1, classification F1, and harmonic F1 scores as functions of confidence threshold for Mask R-CNN models trained on (a) experimental (EXP) data and (b) combined experimental and generated (EXP+GEN) data. All F1 scores are evaluated on the held-out test set of 22 experimental images.}
 \label{fig:confidence-threshold}
\end{figure*}

Figure~\ref{fig:confidence-threshold} shows the detection F1, classification F1, and F1\textsubscript{HM} as functions of the confidence threshold for Mask R-CNN models trained on experimental (EXP) data only and on combined experimental and generated (EXP+GEN) data using 50 original 1024x1024 experimental training images. We observe that the performance metrics of the Mask R-CNN model, specifically defect detection F1 score and classification F1 score, are both influenced by the confidence threshold used during inference. However, their trends differ: the classification F1 score generally increases as a function of the confidence threshold, while the detection F1 score initially increases but eventually decreases at higher thresholds. This divergence arises because higher thresholds filter out low-confidence predictions, which reduces false positives (increasing classification performance), but may also remove true positives (decreasing detection performance). For the EXP model, F1\textsubscript{HM} peaks at a confidence threshold of 0.8, whereas for the EXP+GEN model, the optimal threshold is lower, around 0.6-0.7. We also observe that the optimal threshold may shift depending on the number of training images. In all subsequent analyses, we report the maximum F1\textsubscript{HM} across thresholds as the performance metric for each experiment.

\begin{figure*}[ht]
 \centering
 \includegraphics[width=0.98\textwidth]{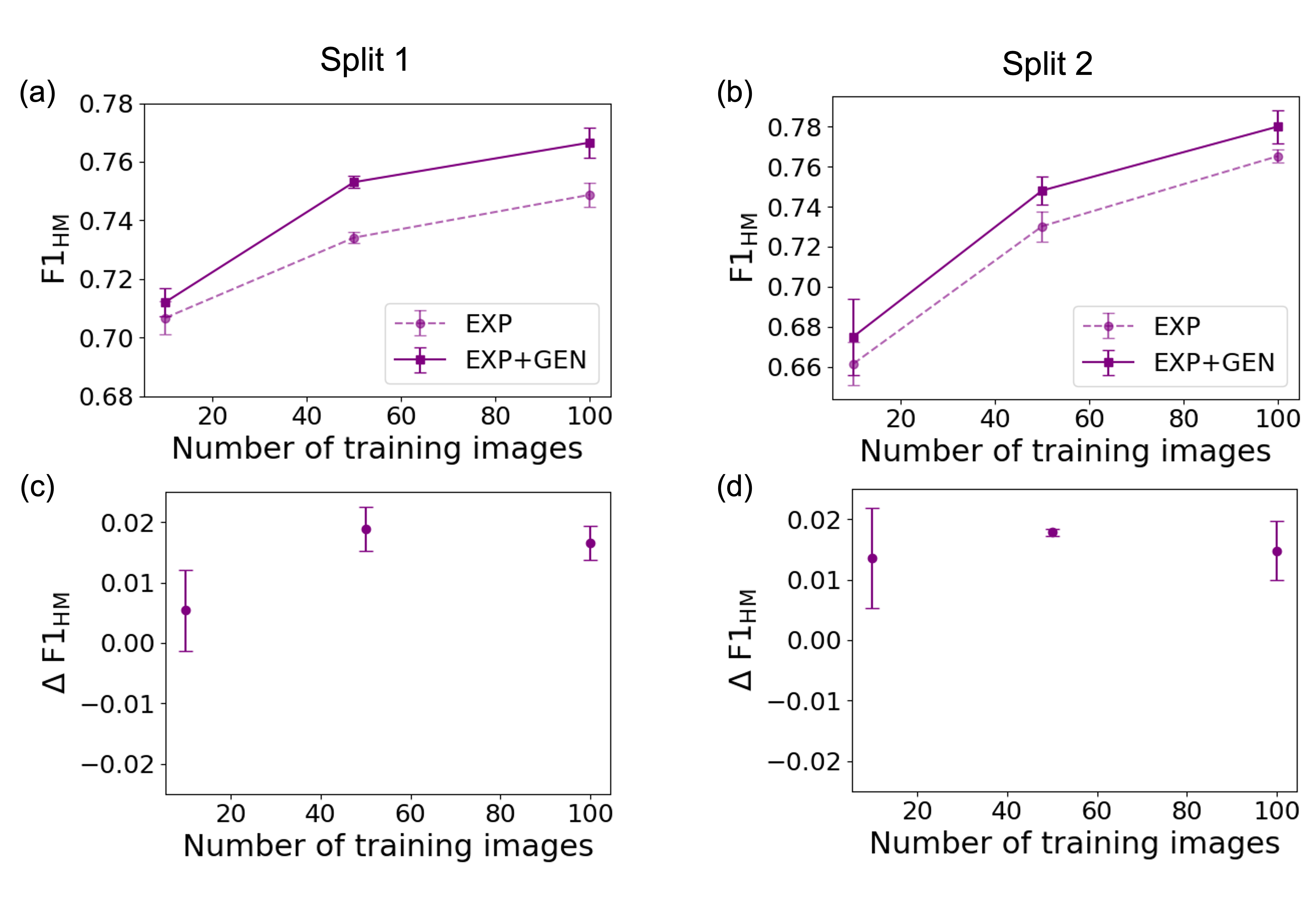}
\caption{Comparison of Mask R-CNN performance across two train/test splits (Split 1 and Split 2). Panels (a) and (b) show the average F1\textsubscript{HM} scores for Mask R-CNN models trained on experimental (EXP) data alone and combined experimental plus generated (EXP+GEN) data, evaluated with 10, 50, and 100 training images. Panels (c) and (d) present the corresponding performance gains ($\Delta$F1\textsubscript{HM}) from generative augmentation. Error bars represent standard deviation across four independent end-to-end runs for each configuration.}
 \label{fig:delta}
\end{figure*}

Figure~\ref{fig:delta} quantitatively shows the Mask R-CNN model performance using F1\textsubscript{HM} and the corresponding performance gain $\Delta$F1\textsubscript{HM} across three training set sizes (10, 50, and 100 images) for Mask R-CNN models trained on experimental (EXP) data alone and a combination of experimental and generated (EXP+GEN) data. Split 1 and Split 2 are two splits that randomly select 22 original images and masks as a testing dataset, and the training set was randomly selected from the remaining images and masks. For each dataset size, we fixed the training/validation split and conducted four independent end-to-end runs to assess the impact of randomness in model training. In each run, the LDM was retrained from scratch using the same EXP data. New synthetic masks were sampled from the real mask distribution, and used to generate corresponding TEM images with the retrained LDM. These synthetic mask-image pairs were then combined with the same experimental dataset to train a new Mask R-CNN model. This full-pipeline process captures the stochastic nature of both the generative model and the instance segmentation task, enabling assessment of the reproducibility and robustness of the model performance. The error bars in Figure~\ref{fig:delta} reflect the standard deviation across these repeated runs.

Figure~\ref{fig:delta}(a) and (b) show that F1\textsubscript{HM}, which combines detection and classification performance, increases with training data size for both EXP and EXP+GEN models. The use of generated data consistently improves performance across all training sizes and across two different data splits. The corresponding performance gain ($\Delta$F1\textsubscript{HM}) is shown in Figure~\ref{fig:delta}(c) and (d), where the improvement is more apparent for larger training sets (50 and 100 images), reaching up to approximately 0.02. This trend indicates that LDM-generated data becomes more beneficial when the generative model is trained with a sufficient number of real images. The improvement is smaller for the 10-image case, likely due to lower generative quality stemming from insufficient data. To assess whether the observed improvements are statistically significant, paired t-tests were performed on four independent runs for each condition in Split 1. The resulting p-values are 0.069 for the 10-image case, 0.0008 for the 50-image case, and 0.0013 for the 100-image case. These results suggest that the performance gains with 50 and 100 real images are statistically significant, while the improvement for 10 images is less conclusive. This observation highlights a significant limitation: while data generation is often motivated as a strategy to overcome the low-data regime, our results suggest that generative models such as LDMs may themselves require a non-trivial amount of real data to produce useful augmentations. This also raises an important practical implication: the LDM pipeline can serve not only as a tool for data augmentation but also as a probe to assess whether additional experimental labeling is likely to yield benefits.

One reasonable method to quantify the the improvement described above is as a percentage of the possible improvement in F1\textsubscript{HM}. Since F1\textsubscript{HM} is bounded by one, the percent of possible improvement is 
$(F1\textsubscript{HM}(EXP+GEN)-F1\textsubscript{HM}(EXP))/(1-F1\textsubscript{HM}(EXP))$. For a typical improvement of 0.01 to 0.02 on F1\textsubscript{HM}=0.70-0.75 this gives a range of about 3-8\%, showing the improvement is quite small but potentially meaningful in applications where one is operating in the low-data limit and acquiring additional experimental data is particularly expensive.

\begin{figure*}[ht]
 \centering
 \includegraphics[width=0.98\textwidth]{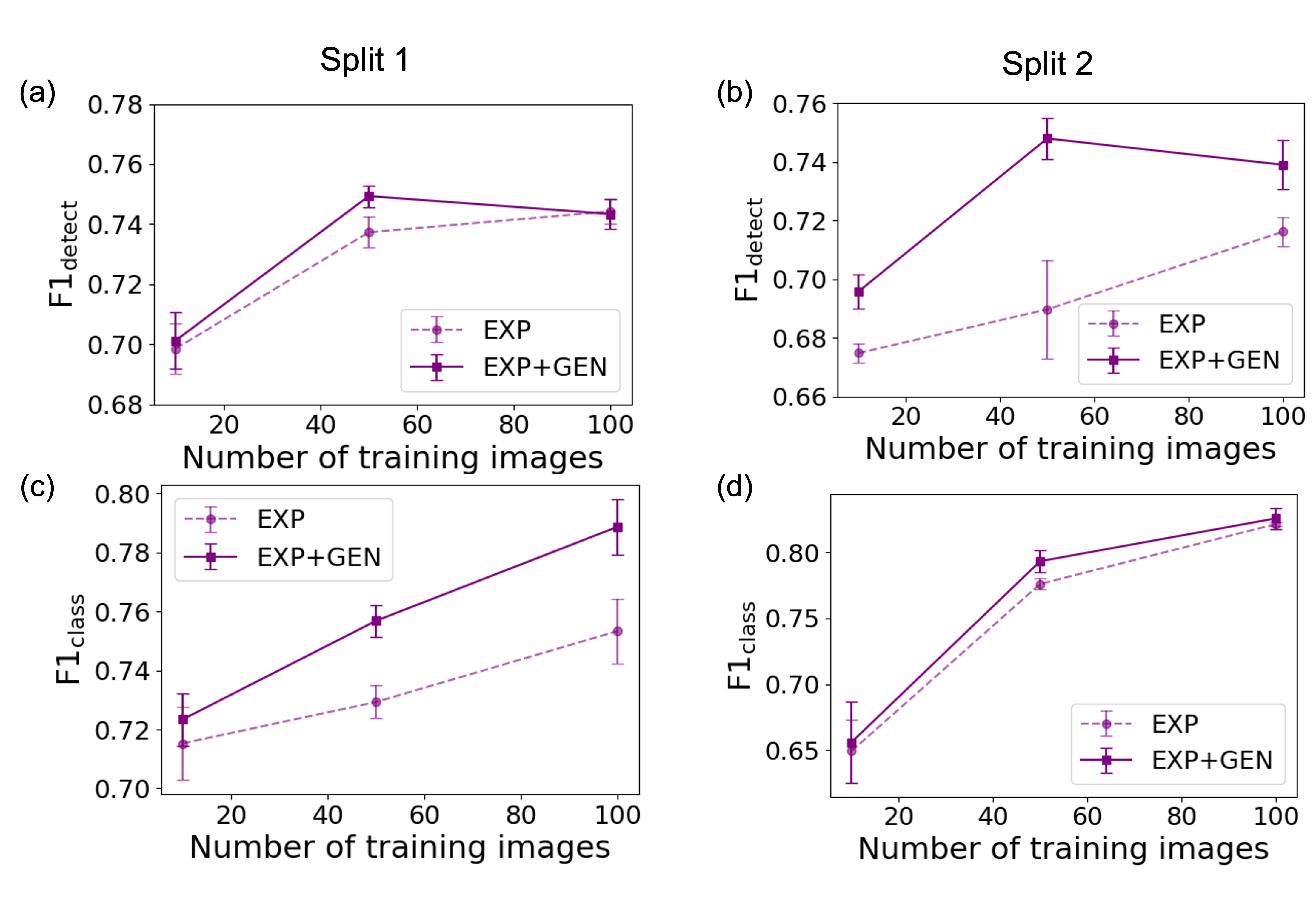}
\caption{Detection and classification performance of Mask R-CNN models trained on experimental (EXP) data alone and on combined experimental and generated (EXP+GEN) data across two train/test splits. Panels (a) and (b) show F1\textsubscript{detect} scores, while panels (c) and (d) show F1\textsubscript{class} scores for Split 1 and Split 2, respectively. Results are reported for training set sizes of 10, 50, and 100 images. Error bars represent standard deviation across four independent end-to-end runs.}
 \label{fig:detect_class_splits}
\end{figure*}

Figure~\ref{fig:detect_class_splits} provides a breakdown of the individual components contributing to the overall performance improvements shown in Figure~\ref{fig:delta}, specifically the detection F1 (F1\textsubscript{detect}) and classification F1 (F1\textsubscript{class}) scores across different training set sizes and two distinct train/test splits. While both splits exhibit similar overall gains in harmonic mean F1 (F1\textsubscript{HM}) when generative data is added, the source of the improvement differs. In Split 1 (panels a and c), the gain is primarily driven by improved classification performance, with minimal change observed in detection scores. The improvement on the classification of each type of defect can be found in the Supplementary Material. In contrast, Split 2 (panels b and d) shows a more substantial increase in detection F1, whereas classification gains are smaller. These results highlight that the benefit of generative augmentation can vary depending on the composition of the train/test split, and improvements in overall performance may arise from either detection or classification, or both, depending on the data partitioning.

Another potential use of LDM-based augmentation is to mitigate class imbalance by preferentially generating underrepresented defect types, such as the  $\langle100\rangle$ loops in this work, as originally found by Jacobs et al.~\cite{RN985} . We briefly explored this idea by attempting to augment the training data by doubling the densities of each specific defect class. However, the preliminary trial did not yield measurable performance improvement. This suggests that simply oversampling a given defect type using our current LDM approach may not be sufficient to address class imbalance, likely because the imbalance pre-exists in the LDM training approach and carried throughout the remainder of the pipeline. Nevertheless, with more careful control over generation quality and defect morphology, targeted class-specific augmentation remains a promising direction for future work.

\section {Summary and Conclusion}

In this work, we proposed a data augmentation framework using a mask-conditioned latent diffusion model (LDM) to synthesize visually consistent scanning transmission electron microscopy (STEM) images with controllable, multi-class defect structures. The synthetic images were conditioned on simulated segmentation masks generated by sampling from geometric distributions learned from experimentally labeled masks. By incorporating these generated image–mask pairs into the training dataset, we aimed to enhance the performance of instance segmentation and classification models.

We quantitatively evaluated the impact of generated data on a Mask R-CNN model trained for both defect detection and multi-class classification across different dataset sizes (10, 50, and 100 images). Our results show that adding 10,000 generated image–mask pairs consistently improves model performance, as measured by the harmonic mean of detection and classification F1 scores (F1\textsubscript{HM}). The average performance gain reaches up to 0.02, mainly for experiments with 50 or 100 real training images. For models trained with only 10 experimental images, however, the improvement is less pronounced, likely due to lower-quality generated images resulting from insufficient training data for the generative model. 
We also examined the separate effects on detection (F1\textsubscript{detect}) and classification (F1\textsubscript{class}) scores. Notably, the source of performance improvement from generative data varies across different train/test splits, with gains arising predominantly from either detection or classification, depending on the split, highlighting the sensitivity of augmentation effects to dataset partitioning.

Overall, our results show that augmenting training data with simulated masks and LDM-generated STEM images can improve the performance of Mask R-CNN models, albeit only slightly, as measured by the harmonic mean of detection and classification F1 scores (F1\textsubscript{HM}). While the improvement is modest (just ~3-8\% of possible improvement available in F1\textsubscript{HM}), it is consistent across training sizes and data splits, demonstrating that generative augmentation with mask-conditioned LDMs can positively influence downstream performance. In the present dataset, however, these gains are likely too small to warrant immediate practical adoption as a standalone solution for significantly improving defect detection and classification. Instead, the results should be interpreted as a proof-of-concept showing that generative augmentation can be beneficial. From a practical TEM workflow perspective, this approach is most relevant in scenarios where an initial dataset has already been collected and a baseline model trained, but further performance optimization is desired while avoiding the cost and effort of additional experimental imaging and annotation. Moreover, the proposed pipeline can serve as a tool to probe whether increasing the effective dataset size is likely to yield meaningful performance improvements, thereby informing decisions about further experimental data collection. 
We focus on $\langle100\rangle$ loops, $\langle111\rangle$ loops, and black dots because they represent technologically important irradiation-induced defects and are supported by well-established annotated datasets; however, the proposed mask-conditioned LDM framework is general and can, in principle, be applied to any defect morphology that can be reliably annotated. Future work may therefore explore extension to other defect types, including voids, stacking fault tetrahedra, and helium bubbles, to evaluate broader applicability under realistic irradiation conditions. It is worth noting that the LDM model was trained on masks annotated by human experts, and thus it is likely to inherit or replicate any bias present in the original annotations. That said, it remains an open and important research question whether and how generative models might subtly shift or exaggerate bias patterns during training. Future studies might also explore additional refinements to the generative methods, e.g., distribution-aware mask sampling,  iterative feedback between detection and generation pipelines. Another area for improvement is in the data, e.g., the use of consensus-labeled ground truth datasets to condition the LDM to avoid the passing of human bias in object labels to the LDM, and, ultimately, the generated images. Additional refinements in the generative methods and the data labeling have the potential to further enable strong performance with limited training data, reducing burdensome data generation and human data labeling needs.

\section*{Author contributions}

Ni L.: Conceptualization; Methodology; Software; Formal analysis; Investigation; Validation; Writing – original draft; Review \& editing Visualization.
Nuohao L.: Methodology; Software; Investigation; Validation; Writing – review \& editing.
Ryan J.: Software; Conceptualization; Resources; Writing – review \& editing.
Dane M.: Conceptualization; Supervision; Writing – review \& editing; Project administration; Funding acquisition.
Ajay A.: Investigation; Review \& editing.
Maciej P.: Conceptualization; Review \& editing.
Kevin F.: Resources; Supervision; Writing – review \& editing.
Izabela S.: Funding acquisition; Supervision; Review \& editing.
\section*{Conflicts of interest}
There are no conflicts to declare.

\section*{Data and code availability}
The experimental dataset analyzed in this study was employed by Jacobs \textit{et al.} and is publicly available on Figshare at \href{https://doi.org/10.6084/m9.figshare.27281400.v1}{https://doi.org/10.6084/m9.figshare.27281400.v1}.  
The code used for data preprocessing, synthetic mask generation, and training of both the latent diffusion model (LDM) and Mask R-CNN model is also accessible on Figureshare (with Github link) at \href{10.6084/m9.figshare.31440037}{10.6084/m9.figshare.31440037}.  
\section*{Acknowledgments}

This work was funded by the Electric Power Research Institute (EPRI) under award number 10012138. Computational resources were provided by the Extreme Science and Engineering Discovery Environment (XSEDE), supported by the National Science Foundation through grant ACI-1548562. Model training and evaluation, including the LDM and Mask R-CNN models, were performed on the Bridges-2 system at the Pittsburgh Supercomputing Center (PSC) under allocation TG-DMR090023, supported by NSF award ACI-1928147. Nuohao L. and Izabela S. acknowledge financial support from  the Department of Energy Basic Energy Science Program (grant number DEFG02–08ER46493)
\bibliographystyle{unsrtnat}
\bibliography{arxiv}

\clearpage
\appendix
\renewcommand{\thesection}{S\arabic{section}}  
\renewcommand{\thesubsection}{S\arabic{section}.\arabic{subsection}}
\renewcommand{\thefigure}{S\arabic{figure}}    
\setcounter{section}{0}
\setcounter{figure}{0}
\setcounter{table}{0}

\end{document}